\pdfoutput=1

\documentclass[11pt]{article}

\usepackage[]{acl}
\usepackage{todonotes}

\usepackage{times}
\usepackage{latexsym}

\usepackage[T1]{fontenc}

\usepackage[utf8]{inputenc}

\usepackage{microtype}

\title{Contextualized Sensorimotor Norms: multi-dimensional measures of sensorimotor strength for ambiguous English words, in context}

\author{Sean Trott \\
  University of California, San Diego \\
  \texttt{sttrott@ucsd.edu} \\\And
  Benjamin Bergen \\
  University of California, San Diego \\
  \texttt{bkbergen@ucsd.edu} \\}

\begin{document}
\maketitle
\begin{abstract}

Most large language models are trained on linguistic input alone, yet humans appear to \textit{ground} their understanding of words in sensorimotor experience. A natural solution is to augment LM representations with human judgments of a word's sensorimotor associations (e.g., the Lancaster Sensorimotor Norms), but this raises another challenge: most words are ambiguous, and judgments of words in isolation fail to account for this multiplicity of meaning (e.g., ``wooden \textit{table}'' vs. ``data \textit{table}''). We attempted to address this problem by building a new lexical resource of contextualized sensorimotor judgments for 112 English words, each rated in four different contexts (448 sentences total). We show that these ratings encode overlapping but distinct information from the Lancaster Sensorimotor Norms, and that they also predict other measures of interest (e.g., relatedness), above and beyond measures derived from BERT. Beyond shedding light on theoretical questions, we suggest that these ratings could be of use as a ``challenge set'' for researchers building grounded language models.

\end{abstract}

\section{Introduction}\label{sec:intro}

Most large language models (LMs) are trained on linguistic input alone. This approach may be fundamentally limited when it comes to language understanding \cite{bender-koller-2020-climbing, bisk-etal-2020-experience, tamari-etal-2020-language}, as the meaning of a word arguably depends on factors beyond which words it co-occurs with. In particular, humans appear to \textit{ground} a word's meaning in a rich network of sensorimotor associations \cite{pulvermuller1999words, bergen_2012_LouderWordsNew, bergen_2008_16EmbodiedConcept, barsalou_1999_PerceptualSymbolSystems, winter2012language, barsalou_2008_GroundedCognition, glenberg2002grounding}. For example, our understanding of the word ``table'' incorporates not just the words that frequently co-occur with ``table'', but also our embodied experience of tables: how they look, how they feel, which parts of our body we use to interact with them, and more. If human-like language understanding depends on grounding words in non-linguistic associations \cite{HARNAD1990335}, then LMs trained on text alone will never reach human levels of understanding \cite{bender-koller-2020-climbing}.

One promising solution is to use human judgments of a word's sensorimotor associations, such as the Lancaster Sensorimotor Norms \cite{lynott2019lancaster} (hereafter LS Norms), to help \textit{ground} LM representations, as well as \textit{evaluate} the extent to which those representations capture sensorimotor properties of word meaning. The LS Norms provide human ratings about the extent to which an isolated word (e.g., ``table'') is strongly associated with various \textit{sensory modalities} (e.g., Vision vs. Touch) and \textit{action effectors} (e.g., Hand/Arm vs. Foot/Leg). Recent work \cite{kennington-2021-enriching, wan-etal-2020-sensorimotor, wan-etal-2020-using} has found that integrating these norms improves the performance of language models on several NLP tasks, such as GLUE \cite{wang-etal-2018-glue} and metaphor detection \cite{wan-etal-2020-using}.

Despite the promise and early success of this approach, it faces a key limitation: resources like the LS Norms typically contain just a single set of judgments for each word. In practice, however, many words are \textit{ambiguous} \cite{rodd2004modelling, haber-poesio-2021-patterns-polysemy}. In English, anywhere from 7\% \cite{rodd2004modelling} to 15\% \cite{trott2020human} of words have multiple, unrelated meanings---and as many as 84\% are polysemous, i.e., they have multiple, related meanings \cite{rodd2004modelling}. For example, the word ``table'' may refer to a piece of furniture or to a database organized into rows and columns. Further, even very similar uses of a word, like ``lemon'', in its fruit-denoting sense,
evoke different sensorimotor associations in different contexts (e.g., ``She peeled the lemon'' vs. ``She put the lemon in the bag'') \cite{yee2016putting, elman2009meaning, trott-etal-2020-construing}. Accordingly, there is evidence that ratings of sensorimotor strength or concreteness can vary considerably depending on whether a word is presented alone or in context \cite{scott2019glasgow}, or as a function of which context a word is presented in \cite{reijnierse2019polysemy}. This suggests that any effort to \textit{ground} words should account for the fact that most words are ambiguous, with dynamic, context-sensitive meanings subject to construal. Further, attempts to \textit{evaluate} grounded language models must consider not only how well they capture the sensorimotor properties of a word in isolation, but also how successfully they capture context-dependent variation in a word's sensorimotor profile.

In Section \ref{sec:related}, we first describe related resources, as well as work on grounding large LMs using psycholinguistic resources and multimodal input. In Section \ref{sec:csn}, we introduce the Contextualized Sensorimotor Norms (CS Norms), a dataset of sensorimotor judgments about ambiguous words in context. In Section \ref{sec:characterization}, we provide descriptive statistics about the CS Norms, as well as comparisons to other factors such as the \textit{dominance} of a particular sense. In Section \ref{sec:utility}, we show that a metric derived from the CS Norms---the Sensorimotor Distance between two contexts of use---improves our ability to predict contextualized relatedness judgments, above and beyond a similar metric derived from BERT \cite{devlin-etal-2019-bert}. Finally, in Section \ref{sec:discussion}, we discuss limitations of these norms, as well as avenues for future research.

\section{Related Resources}\label{sec:related}

\subsection{Related Resources}

There are a number of existing lexical resources with information about the concreteness or sensorimotor strength of words \cite{coltheart1981mrc, brysbaert2014concreteness}. For example, the Brysbaert concreteness norms contain concreteness judgments for approximately 37,000 English words \cite{brysbaert2014concreteness}; concreteness ratings have also been collected for Dutch \cite{brysbaert2014norms}, Croatian \cite{coso2019affective}, and more.

Judgments of concreteness or overall sensorimotor strength are limited in that they do not account for which sensorimotor features are particularly salient. More recently, researchers have collected ratings about multiple semantic features for each word, including its sensorimotor associations \cite{lynott2019lancaster}, as well as even more fine-grained judgments within each modality (e.g., for Vision, whether the referent is Fast or Slow; for Touch, whether it is Hot or Cold) \cite{binder2016toward}. Of these, the largest dataset is the Lancaster Sensorimotor Norms \cite{lynott2019lancaster}, which includes 11-dimensional judgments for about 40,000 English words. This approach has been extended to other languages, such as French \cite{miceli2021perceptual} and Dutch \cite{speed2021dutch}. Again, in each case, the words were presented without context.

Finally, several datasets have collected concreteness judgments about words in context \cite{scott2019glasgow, reijnierse2019polysemy}. However, to our knowledge, no dataset includes judgments about \textit{which} sensorimotor features are particularly salient in different linguistic contexts.

\subsection{Grounding LMs with Psycholinguistic Resources}

Recent work in NLP has begun to incorporate these psycholinguistic resources. One approach attempts to predict these judgments about concreteness or salient sensorimotor features from LM representations, with varying degrees of success \cite{thompson2018automatic, turton-etal-2020-extrapolating, chersoni-etal-2020-automatic, utsumi2020exploring}. 
Another approach uses sensorimotor features to augment the ability of an LM on an applied task, such as the GLUE benchmark \cite{kennington-2021-enriching} or metaphor detection \cite{wan-etal-2020-sensorimotor}. Ass mentioned in Section \ref{sec:intro}, these experiments are limited in that the sensorimotor features themselves were obtained for words in isolation.

\subsection{Grounding LMs with Multimodal Input}

An alternative approach is to ground LM representations more directly in multimodal input. Most of this work has emphasized the visual modality, linking words to static images \cite{kiros-etal-2018-illustrative, su2020vlbert} or video \cite{zellers2021merlot}. This paradigm shows considerable promise, though it is naturally limited by resource constraints; obtaining reliable multimodal data and aligning it to language can be both time-consuming and costly.

\subsection{Summary}

There is considerable interest in \textit{grounding} among both psycholinguists and NLP practitioners. To that end, psycholinguists have developed large linguistic resources, which some NLP researchers have used to improve LMs. 

Still, one limitation of the majority of existing resources is that they do not contain judmgents about different sensorimotor features for words in different contexts. Because most words are ambiguous, this makes it difficult to know which meaning the sensorimotor judgments reflect, which in turn reduces the precision and utility of these resources.

\section{Contextualized Sensorimotor Norms}\label{sec:csn}

Our primary goal was to collect sensorimotor judgments about ambiguous words, appearing in controlled sentential contexts. We used sentences from the RAW-C (Relatedness of Ambiguous Words--in Context) dataset \cite{trott-bergen-2021-raw}. RAW-C contains relatedness judgments for 672 English sentence pairs, each containing the same target word (e.g., ``bat'') in either the same meaning (e.g., ``furry bat'' vs. ``fruit bat'') or different meaning (e.g., ``furry bat'' vs. ``wooden bat''); it also contains dominance judgments about the relative salience of each meaning. There were 448 unique sentences in total (112 target words, with 4 sentences each).

We collected judgments about the sensorimotor associations evoked by the target word in each of the 448 sentences. This provided a more direct analogue to the Lancaster Sensorimotor Norms \cite{lynott2019lancaster}, in which participants observed a particular lexical item (e.g., ``bat'') and provided ratings about its associated sensory modalities (e.g., Vision) or action effectors (e.g., Hand/Arm). 

\subsection{Participants}

Our goal was to collect a minimum of 10 judgments per sentence. Thus, we recruited participants until each sentence had at least 10 observations, after applying the exclusion criteria.

A total of 377 participants were recruited through UC San Diego's undergraduate subject pool for Psychology, Cognitive Science, and Linguistics students. Participants received class credit for participation. After excluding non-native speakers of English, participants who failed to pass the bot checks, and participants whose inter-annotator agreement score was sufficiently low (see Section \ref{sec:agreement} below), we were left with 283 participants. Of these, 223 identified as female (47 male, 8 non-binary, and 5 preferred not to answer). The mean self-reported age was 20.4 (median = 20, SD = 2.98), and ranged from 18 to 43.

\subsection{Procedure}

We adapted the procedure directly from \citet{lynott2019lancaster}, with the main modification being that participants now saw words in sentential contexts.
Participants were randomly assigned to one of two Judgment Types: 1) Perception, in which they provided ratings about a word's associated sensory modalities (Vision, Hearing, Touch, Interoception, Smell, and Taste); and 2) Action, in which they rated a word's associated action effectors (Hand/Arm, Foot/Leg, Mouth/Throat, Head, and Torso). In total, 132 participants were assigned to the Perception Judgment Type, and 151 were assigned to the Action Judgment Type.

After giving consent, participants answered two bot check questions. They were then told that they would read a series of sentences, each containing a bolded word (e.g., ``It was a wooden \textbf{table}''), and that their task was to rate the degree to which they experienced the concept denoted by that word with either six sensory modalities (in the Perception Judgment Type) or five action effectors (in the Action Judgment Type). Ratings ranged from 0 (not at all experienced with that sense/effector) to 5 (experienced greatly with that sense/effector). 

Each participant rated approximately 60 sentences overall, randomly sampled from the set of 448 sentences. No participant saw the same target word twice. On each trial, the sentence was displayed at the top of the page, with the target word bolded. Underneath the sentence, the instructions read: ``To what extent do you experience WORD:'' (for Perception) or ``To what extent do you experience WORD by performing an action with the:'' (for Action), where ``WORD'' was replaced with the target word. Underneath the instructions were six (for Perception) or five (for Action) rating scales, corresponding to each possible sensory modality or action effector. For the Action Judgment Type, the scale was accompanied by a labeled diagram of the body, as in \citet{lynott2019lancaster}.

To reach the target of 10 respondents to each word in both Action and Perception tasks, we collected data in two stages. In the first stage (Group 1), participants were randomly assigned to either the Perception or Action Judgment Types, and the sentences they observed were randomly sampled from the set of possible sentences for each word. After we had collected responses from 264 participants in this way, there were still a number of sentences that had very few observations, simply by chance---as well as many with more than ten observations. Thus, in the second stage (Group 2), participants were assigned a mix of Low-N (sentences with fewer than 10 ratings) and High-N (sentences with 10 or more ratings) items. The goal was to speed data collection; to control for potential differences across groups, we compared their distributions of inter-annotator agreement scores, and found no evidence that the different data collection procedures induced different response behavior (see Section \ref{sec:agreement}).

Finally, after providing ratings, participants reported their self-identified gender and age, as well as whether or not they were a native speaker of English.

The data collection was conducted online using JsPsych \cite{de2015jspsych}. 

\subsection{Inter-Annotator Agreement}\label{sec:agreement}

We sought to establish the degree to which different participants agreed about their ratings for each sentence, both to characterize the dataset and to exclude participants whose ratings diverged substantively from the rest of the sample. Following past work \cite{trott-bergen-2021-raw}, we used a leave-one-out scheme: for each participant, we computed the Spearman's rank correlation between that participant's responses and the mean ratings for those items from the rest of the sample (excluding the participant's ratings). 

Importantly, we did this in two stages. First, we computed the distribution of agreement scores for the 264 participants in Group 1, i.e., the participants for whom each sentence was truly randomly sampled from the set of 448 sentences. Based on this distribution of inter-annotator agreement scores, we excluded a total of 18 participants, whose scores were more than two standard deviations below the mean for that Judgment Type. Among the final set of 246 participants in this group, the mean inter-annotator rank correlation was 0.47 for Action judgments (SD = 0.1) and 0.64 for Perception judgments (SD = 0.11). 

Then we considered the 39 participants from Group 2, who provided ratings for a restricted set of sentences, i.e., sentences which either had below 10 judgments from Group 1 (low-N) or had more than 10 judgments from Group 1 (high-N). For each participant in Group 2, we compared the ratings for the high-N items to the mean response for those items among Group 1. After excluding participants with low inter-annotator agreement, we were left with a total of 37 participants in Group 2. The mean rank correlation was 0.5 for Action Judgments (SD = 0.11) and 0.64 for Perception judgments (SD = 0.1). 

Finally, we combined the set of inter-annotator agreement scores from both groups, and constructed a linear regression with Rank Correlation as the dependent variable, and main effects of Judgment Type (Action vs. Perception) and Group (Group 1 vs. Group 2), as well as their interaction. There was no significant difference in agreement across groups ($p > .1$), but agreement was significantly higher for Perception ratings than Action ratings [$\beta = 0.17, SE = 0.01, p < .001$].

\subsection{Creating the Norms}

Once we had obtained a minimum of ten ratings per sentence (per judgment type), we averaged across these ratings to produce a mean and standard deviation for each dimension. For example, the sentence ``He saw the furry \textbf{bat}'' would contain the mean (and standard deviation) of judgments about the salience of each sensorimotor feature.\footnote{The norms (along with the individualized responses, analysis code, and a Data Sheet) can be found on GitHub: \url{https://github.com/seantrott/cs_norms}.}

\section{Characterizing the Contextualized Sensorimotor Norms}\label{sec:characterization}

Our first goal was to characterize the Contextualized Sensorimotor Norms (CS Norms). The norms provide an 11-dimensional vector for each sentential context in which a word appears: the mean sensorimotor strength for 11 dimensions (6 sensory modalities, and 5 action effectors) for a target word in a given context.

\begin{figure}
    \centering
    \includegraphics[width=7cm]{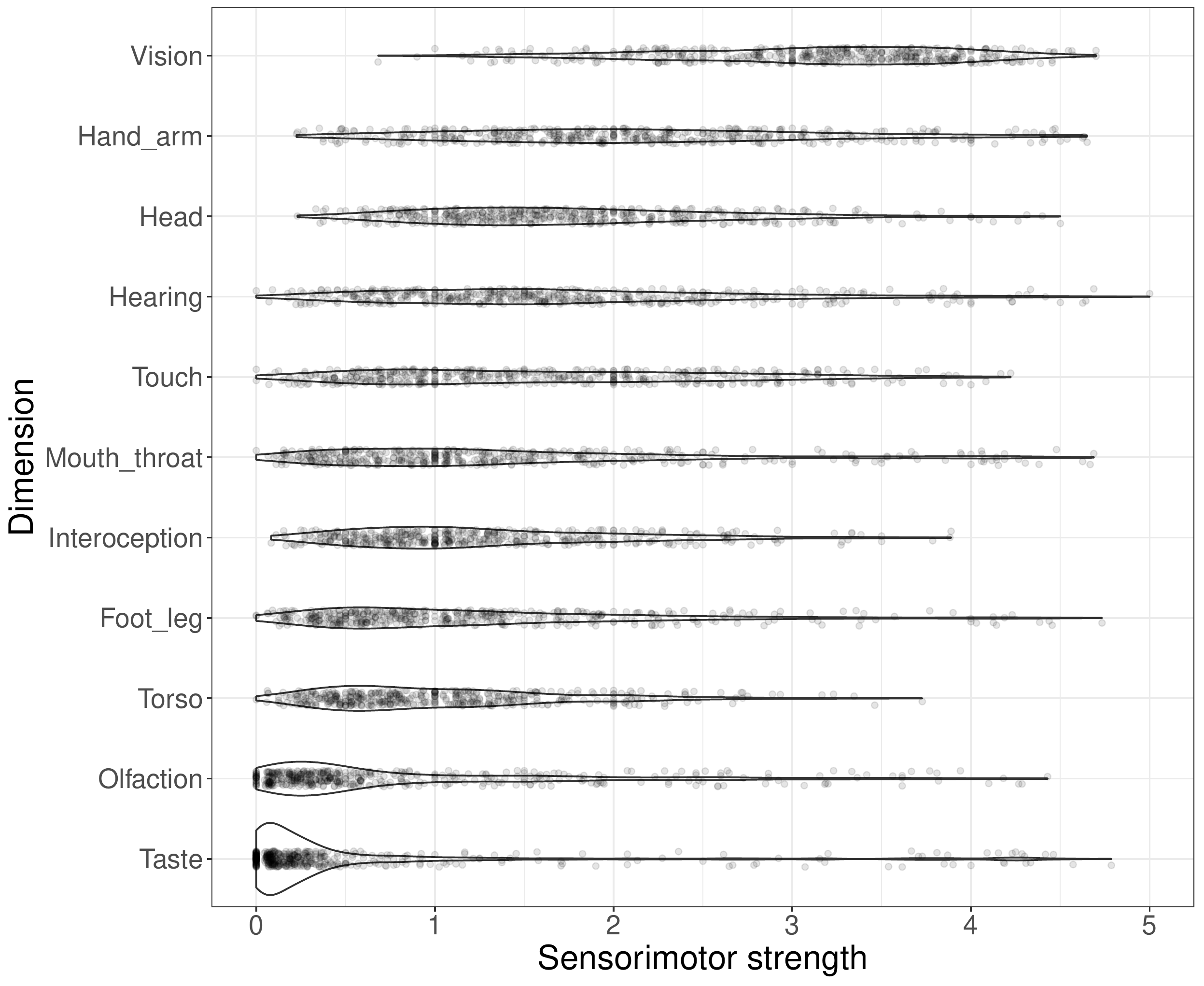}
    \caption{Distribution of mean sensorimotor strength judgments for each dimension. As in past work \cite{lynott2019lancaster}, judgments were highest for the Vision dimension, and lowest for Olfaction and Taste.}
    \label{fig:distributions}
\end{figure}

\subsection{Comparing Sensorimotor Dimensions}

As a first step, we visualized the distribution of sensorimotor judgments for each dimension (see Figure  \ref{fig:distributions}). Consistent with the original LS Norms \cite{lynott2019lancaster} and work on the English lexicon more generally \cite{majid2020human}, judgments tended to be highest for the Vision dimension, and lowest for Olfaction and Taste.

\begin{figure}
    \centering
    \includegraphics[width=7cm]{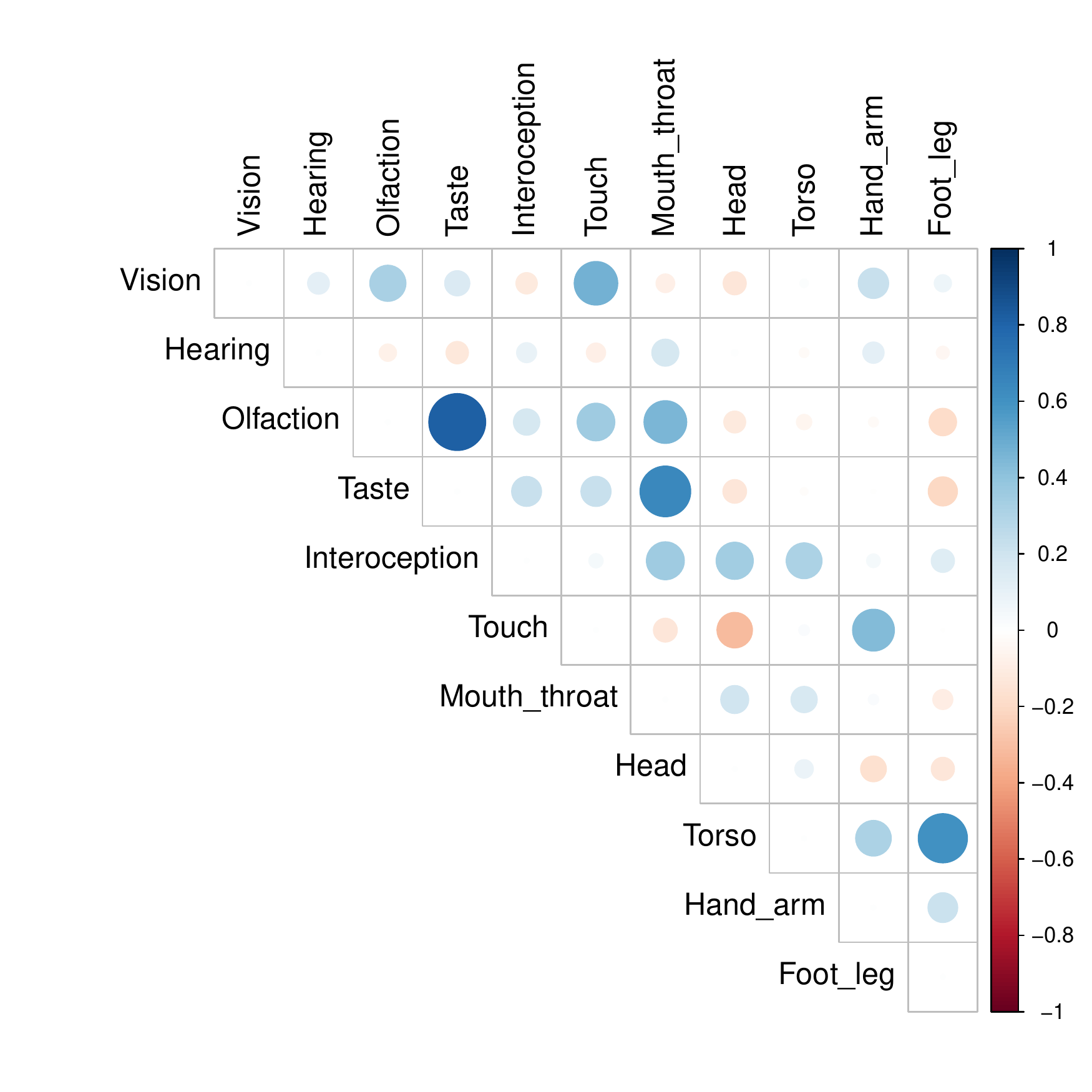}
    \caption{Pearson's correlation coefficients between the sensorimotor strength of each feature.}
    \label{fig:corrs}
\end{figure}

We then asked which dimensions were correlated with which other dimensions. Consistent with past work \cite{lynott2019lancaster}, we found particularly strong positive correlations between Olfaction and Taste, as well as Foot/Leg and Torso; we also found a strong positive correlation between Taste and Mouth/Throat (see Figure \ref{fig:corrs}). 

\subsection{Variance Across Contexts}\label{subsec:compare_to_lancaster}

A key motivation for the CS Norms was to account for potential variation within each word in terms of which sensorimotor features were most salient across distinct sentential contexts.

We first quantified this variation by normalizing the sensorimotor features for each context of use to the mean norms for that word from the LS Norms.\footnote{See Section \ref{sec:dominance2} for further quantification of this variability.} For example, the LS norms have a single 11-dimensional vector for the word ``market''; for each of the four sentential contexts in which ``market'' appeared, we calculated the difference in mean ratings across our norms and the LS Norms. This provides an estimate of the degree to which the human judgments were impacted by the sentential context, as opposed to a representation of the word's meaning in isolation (as in the LS Norms).

\begin{figure}
    \centering
    \includegraphics[width=7cm]{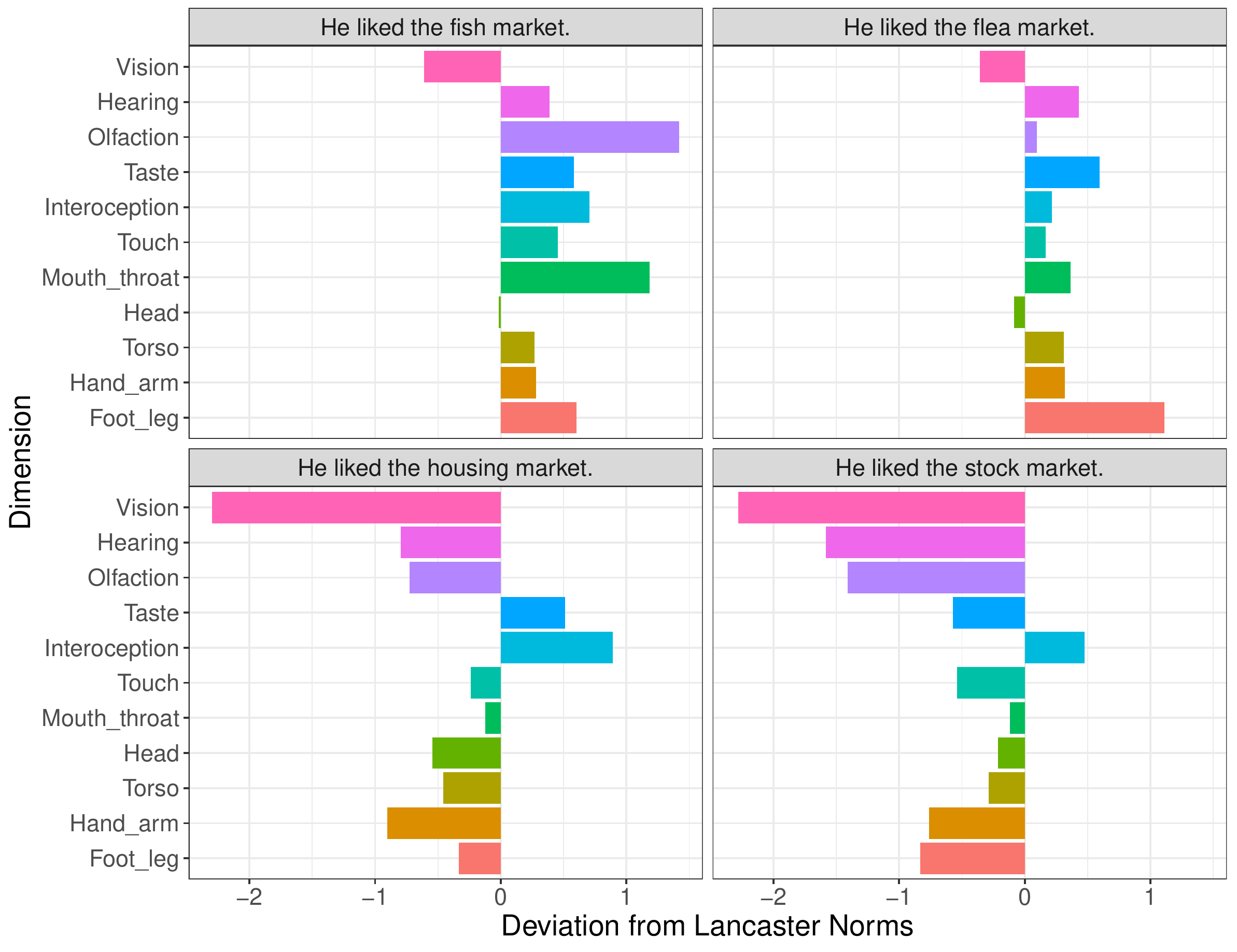}
    \caption{Deviation between the contextualized sensorimotor norms and the Lancaster Sensorimotor Norms for for the word ``market'', faceted by the distinct sentential context in which ``market'' appeared.}
    \label{fig:market}
\end{figure}

Figure \ref{fig:market} depicts these deviations from the LS Norms for a specific word, ``market''. This word was chosen because it displayed particularly high variation in its overall sensorimotor strength across contexts. The deviations from the LS Norms appear to track the two senses of the word being profiled. The two sentences corresponding to the \textit{location} sense of ``market'' (i.e., ``fish market'' and ``flea market'') appeared to be closer to the LS Norms (i.e., the deviations were smaller on average);  the notable exceptions were the \textit{Olfactory} and \textit{Mouth/Throat} dimensions for the ``fish market'' context, and the \textit{Foot/Leg} dimension for the ``flea market'' context. In contrast, the sentences corresponding to the the  \textit{financial} sense of ``market'' (i.e., ``housing market'' and ``stock market'') were considerably lower in sensorimotor strength across almost all dimensions, especially \textit{Vision}. This makes sense, given that this meaning is more metaphorical or abstract than the \textit{location} meaning of ``market'': apart from representations of their performance, neither housing markets nor stock markets can be visually perceived in the way that fish markets and flea markets can.

\subsection{Sense Dominance and Deviation from the Lancaster Norms}\label{sec:dominance2}

One well-documented property of ambiguous words is that their multiple meanings are not always balanced: one sense is sometimes more cognitively salient than the other. This is called \textit{sense dominance}. The degree of dominance is known to play an important role in the processing of ambiguous words, particularly for homonyms: empirical evidence suggests that comprehenders almost always activate the more dominant sense of a homonym, even when the linguistic context supports the subordinate meaning \cite{rayner1994effects, binder1998contextual, duffy1988lexical}. We used the measure of Sense Dominance in the RAW-C dataset to answer two additional questions about how and why the CS Norms deviate from the decontextualized LS Norms.

First, do the decontextualized LS Norms primarily reflect the more dominant meaning of an ambiguous word? To answer this, we calculated the cosine distance between the decontextualized LS Norm for each word and the sensorimotor norms for that same word in context. We called this Distance to Lancaster. Using the \textit{lme4} package \cite{bates2007lme4} in R, we fit a linear mixed effects model predicting Distance to Lancaster, with Dominance as a fixed effect (and random intercepts for words); this model explained more variance than a model omitting only Dominance $[\chi^2(1)=10.93, p = .001]$. More dominant senses were closer to the decontextualized norm on average $[\beta = -0.01, SE = 0.003, p = .001]$, consistent with the prediction that the LS Norms are more influenced by properties of the dominant sense.

Second, do more dominant senses have stronger or weaker sensorimotor ratings, on average, than the decontextualized rating for that word? For each sentential context, we computed the average difference between each dimension and the corresponding LS Norms, such that a positive value reflects a \textit{more} concrete context. A model predicting this measure was significantly improved by the addition of Dominance $[\chi^2(1)=38.24, p < .001]$. Dominant senses tended to have stronger sensorimotor ratings, on average, than the decontextualized ratings for that same word $[\beta = 0.09, SE = 0.01, p = .001]$.

\subsection{Sense Dominance and Sensorimotor Strength}\label{sec:dominance1}

We also sought to replicate previous work suggesting that more dominant meanings tend to be more concrete \cite{gilhooly1980meaning}. Following \citet{lynott2019lancaster}, we created a composite variable called Contextualized Sensorimotor Strength, which measured the maximum strength across the 11 sensorimotor features for each context of use. Then, we built a linear mixed effects model with Dominance as a dependent variable, fixed effects of Contextualized Sensorimotor Strength, random intercepts for each word, and two covariates reflecting the decontextualized sensorimotor strength for each \textit{word} (i.e., from the Lancaster Sensorimotor Norms dataset). The full model explained significantly more variance than the same model omitting only  Contextualized Sensorimotor Strength  $[{\chi}{^2}(1)=18.38, p  < .001]$. Consistent with past work \cite{gilhooly1980meaning}, contexts of use with higher sensorimotor strength were also rated as more dominant $[\beta = 0.26, SE = 0.06, p < .001]$. (Of course, this finding does not explain \textit{why} more concrete meanings are more dominant than meanings with less sensorimotor strength; it could be driven by correlations with meaning frequency or even age of acquisition \cite{gilhooly1980meaning}.)

\subsection{Sensorimotor Distance}\label{subsec:sm_distance}

Another question concerns the relationship \textit{between} contexts of use. Each context of use for a given wordform is associated with its own sensorimotor norms, i.e., the mean ratings for each sensorimotor dimension for a given context; because of this, the similarity or dissimilarity between these contexts can be quantified by calculating the cosine distance between these vectors \cite{wingfield2021sensorimotor}. 
Thus, we calculated the cosine distance---referred to here as the Sensorimotor Distance---between the vectors corresponding to each sentence pair for each word (672 sentence pairs total). Larger distances reflect more dissimilar contexts of use, while smaller distances reflect more similar contexts. Note that this includes comparisons between contexts corresponding to the same sense and those corresponding to different senses.

We then asked whether Sensorimotor Distance was correlated with other psychologically relevant features, such as whether the two contexts of use corresponded to the same sense or different senses (i.e., Sense Boundary). Based on the preliminary findings in Section \ref{subsec:compare_to_lancaster}, we predicted that different sense uses would have less similar sensorimotor features.

\begin{figure}
    \centering
    \includegraphics[width=7cm]{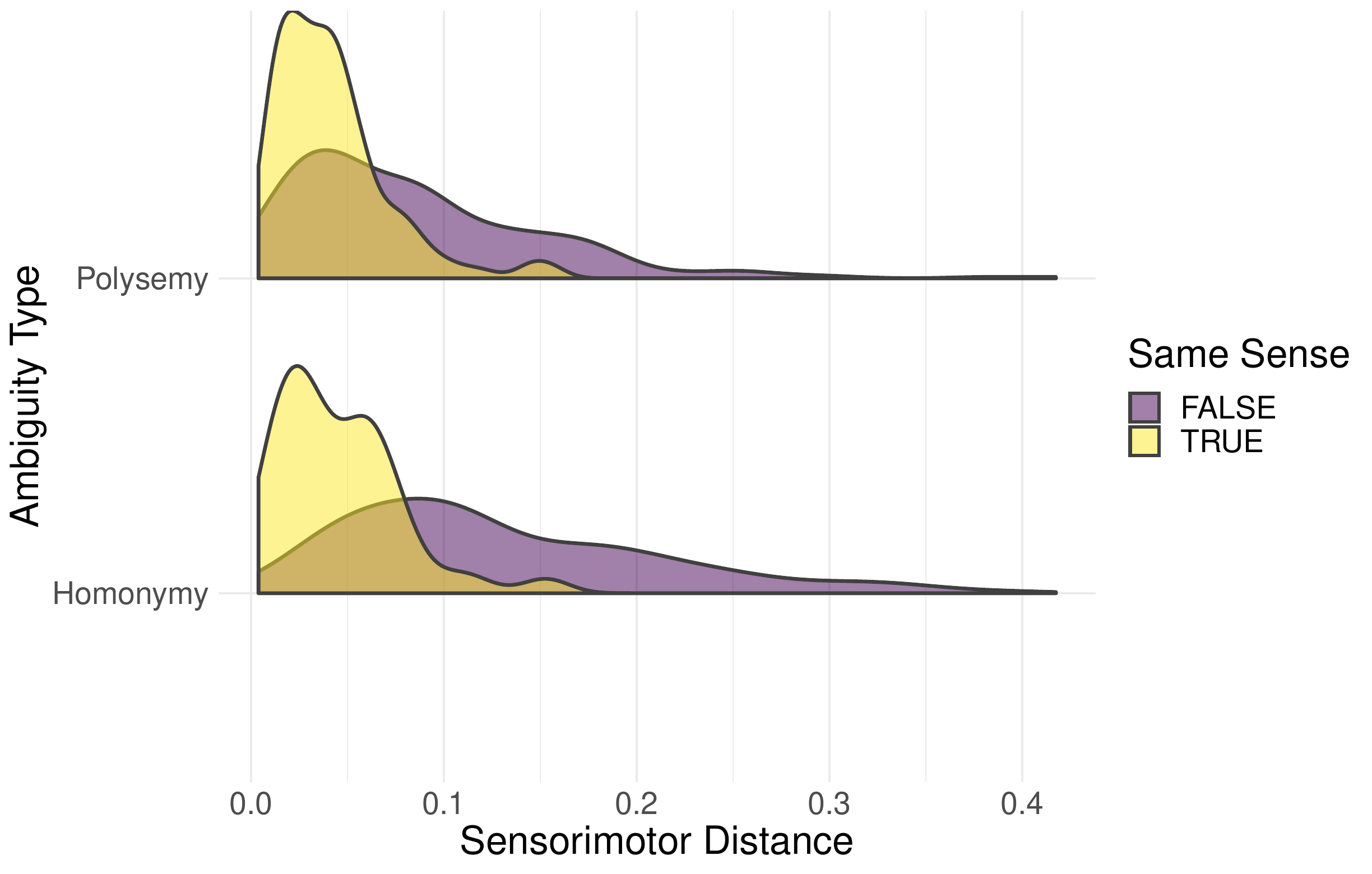}
    \caption{Distribution of sensorimotor distances as a function of same/different sense, as well as the type of ambiguity. Same sense uses have more similar sensorimotor associations than different sense uses.}
    \label{fig:same_diff}
\end{figure}

Indeed, as depicted in Figure \ref{fig:same_diff}, Sensorimotor Distance was considerably larger for Different Sense than Same Sense contexts. The addition of Sense Boundary to a mixed effects model predicting Sensorimotor Distance improved model fit beyond a model with only Distributional Distance and Ambiguity Type (and random intercepts for words) $[\chi^2(1)=34.86, p < .001]$. This is also consistent with Figure \ref{fig:market}, in which the two \textit{location} senses of ``market'' were more similar to each other than either was to the two \textit{financial} senses.

\section{Predictive Utility}\label{sec:utility}

We were also interested in the predictive utility of the information provided by the CS Norms, above and beyond other commonly used factors. To what extent do these contextualized ratings encode information that large language models (e.g., BERT) or decontextualized sensorimotor norms (e.g., LS Norms) fail to capture? 

We sought to predict the \textit{relatedness} of sentence pairs. RAW-C contains judgments of sense relatedness for each unique sentence pair within each of the 112 words, with a total of 672 sentence pairs \cite{trott-bergen-2021-raw}. It is also annotated for whether the two contexts of use correspond to the same or different sense (Sense Boundary), and whether the relationship type is one of homonymy or polysemy (Ambiguity Type). Additionally, past work \cite{trott-bergen-2021-raw} found that relatedness was negatively correlated with the cosine distance between BERT's contextualized embeddings for the target word in each sentence; here, we call this measure the \textit{Distributional Distance}. 

\begin{figure}
    \centering
    \includegraphics[width=7cm]{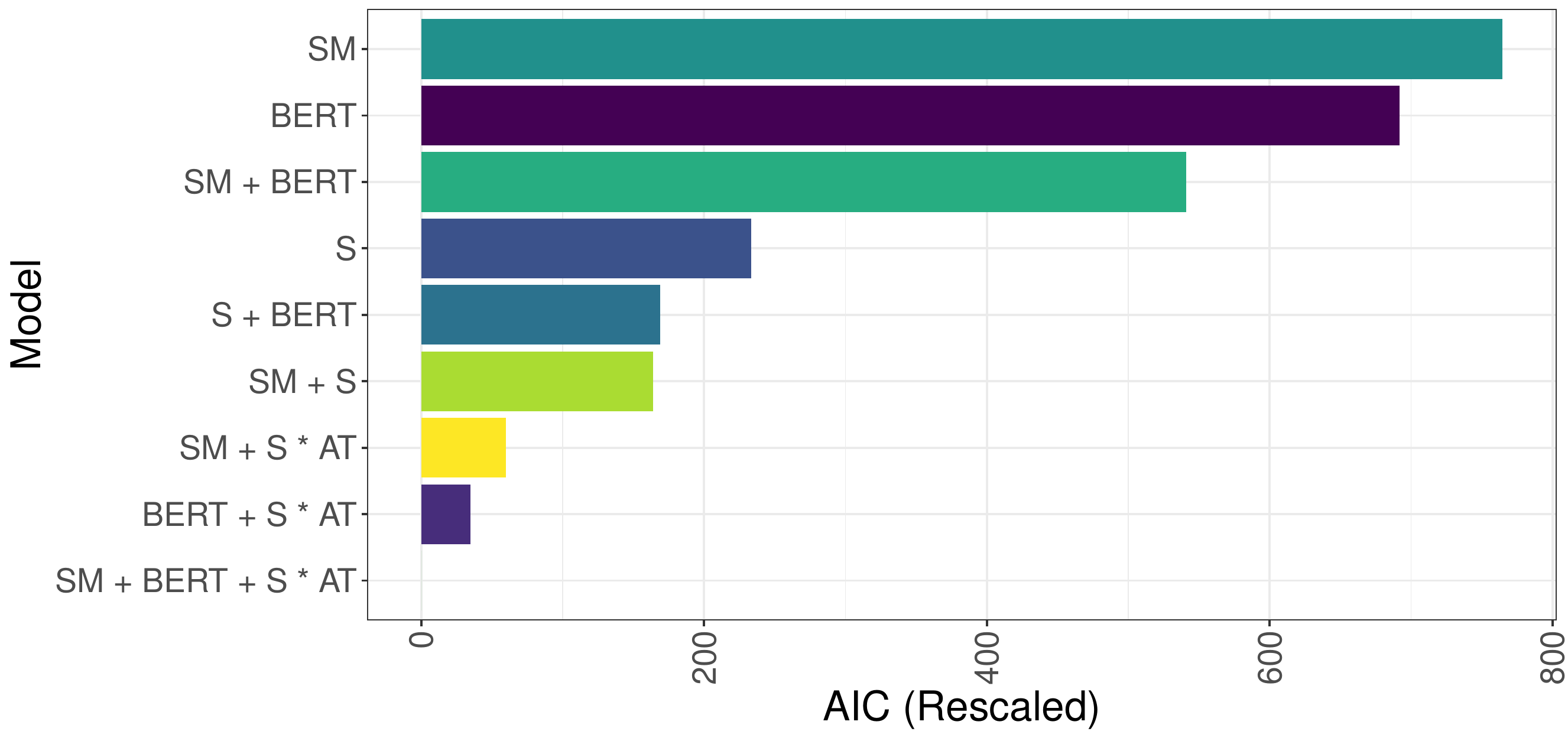}
    \caption{Rescaled AIC values for models predicting Relatedness using an assortment of factors: Sense Boundary (S), Ambiguity Type (AT), Distributional Distance (BERT), and Sensorimotor Distance (SM). A lower AIC score corresponds to better model fit.}
    \label{fig:aic}
\end{figure}

We asked whether a linear mixed effects model equipped with those previous factors (Distributional Distance,\footnote{Distributional Distance was calculated by taking the cosine distance between the final layers of BERT's contextualized embeddings for the target word in each sentence, using the \texttt{bert-embedding} package (\url{https://pypi.org/project/bert-embedding/}).} Sense Boundary, Ambiguity Type, and their interaction, as well as random intercepts for words) could be improved by the addition of Sensorimotor Distance (see Section \ref{subsec:sm_distance}). Indeed,  Sensorimotor Distance significantly improved model fit $[{\chi}{^2}(1)=36.74, p  < .001]$. As expected, Sensorimotor Distance was negatively associated with Relatedness $[\beta = -1.81, SE = 0.22, p < .001]$: words with more dissimilar sensorimotor vectors were rated as less related, on average (see also \ref{fig:aic}).

We then compared the Akaike Information Criterion, or AIC, of a number of different models predicting Relatedness. The models were constructed to probe the explanatory value of the distance measures, as well as the categorical condition variables (e.g., Sense Boundary). Each statistical model under consideration contained at least one of the following variables: Sensorimotor Distance (SM), BERT Distance (BERT), Sense Boundary (S), Ambiguity Type (AT), and an interaction between Sense Boundary and Ambiguity Type (S * AT). 

Crucially, the inclusion of Sensorimotor Distance consistently improved model fit. In other words, the CS Norms appear to capture information that is at least partially independent from the information encoded by factors such as BERT Distance, Sense Boundary, and Ambiguity Type. Of course, it is also important to note that Sense Boundary was by far the best predictor of Relatedness, suggesting that neither distributional similarity nor sensorimotor similarity are sufficient to account for the possible effect of categorical sense representations (see Figure \ref{fig:aic}).

\subsection{Comparison to Lancaster Sensorimotor Norms}

We also constructed a ``baseline'' model using the decontextualized LS Norms. We first identified the disambiguating word across each pair of sentence contexts (e.g., ``\textit{furry} bat'' vs. ``\textit{wooden} bat''); then, we calculated the cosine distance between the decontextualized LS Norms corresponding to each of those words (e.g., between ``furry'' and ``wooden''). We called this measure the Decontextualized Distance. Because not every disambiguating word was included in the LS Norms (e.g., ``double-sided''), we excluded some pairs from the analysis, resulting in 576 pairs total. 

Crucially, a linear mixed effects model including both Contextualized Sensorimotor Distance and Decontextualized Distance explained more variance than a model omitting only Contextualized Sensorimotor Distance $[\chi^2(1)=142.07, p < .001]$. This indicates that the contextualized measure (i.e., from the CS Norms) encodes additional information beyond what can be inferred by simply identifying the sensorimotor properties of other words in the context.

\section{Discussion}\label{sec:discussion}

Embodied experience appears to be crucial for how humans learn and understand language \cite{bergen_2012_LouderWordsNew, pulvermuller1999words, barsalou_1999_PerceptualSymbolSystems}, yet most large language models (LMs) are exposed to linguistic input alone \cite{bender-koller-2020-climbing}. One solution is to \textit{augment} and \textit{evaluate} LM representations using psycholinguistic resources, such as human judgments of the sensorimotor features associated with a word \cite{lynott2019lancaster}. However, this approach must also contend with the challenge of lexical ambiguity. Words mean different things in different contexts \cite{rodd2004modelling, trott-etal-2020-construing}, yet many lexical resources collect judgments about words in isolation.

We attempted to address this challenge by collecting judgments about the salience of various sensory modalities (e.g., \textit{Vision}) and action effectors (e.g., \textit{Torso}) for the same English word, in distinct sentential contexts (e.g., ``flea \textit{market}'' vs. ``housing \textit{market}''). We called this dataset the Contextualized Sensorimotor Norms (CS Norms).

These contextualized norms capture variance in sensorimotor associations beyond the information already provided by the Lancaster Sensorimotor Norms (Figure \ref{fig:market}). We also replicated past work \cite{gilhooly1980meaning} suggesting that the psychological dominance of a meaning is correlated with its sensorimotor strength. Third, we found that the sensorimotor distance between contexts of use was correlated with the existence of sense boundary (see Figure \ref{fig:same_diff}). In Section \ref{sec:utility}, we demonstrated the predictive utility of the CS Norms above and beyond large LMs such as BERT and the decontextualized LS Norms (Figure \ref{fig:aic}). 

Beyond its use for NLP applications, the CS Norms could also be used to address questions of theoretical interest, such as the relative contribution of different sources of information (e.g., distributional vs. sensorimotor associations) to semantic representations in the mental lexicon \cite{andrews2014reconciling, davis2021building}, as well as a role for discrete, symbolic representations (e.g., sense boundaries). For example, as with the LS Norms, researchers could use sentences from this dataset as stimuli in behavioral or neuroscientific experiments. 

\subsection{Limitations}

This dataset is not without limitations.

First, it is restricted in size and breadth: 448 sentences (112 words, with 4 sentences each), in English only. In contrast, the Lancaster Sensorimotor Norms contain judgments of almost 40,000 English words \cite{lynott2019lancaster}, and have now been extended to French \cite{miceli2021perceptual}, Dutch \cite{speed2021dutch}, and more. Having demonstrated the utility of the CS Norms on a small subset of English words, one obvious direction for future research would be to expand this dataset---including more words, more senses and sentences per word, a wider variety of sentences (i.e., both experimentally controlled and naturalistic sentences), and additional languages. Similarly, existing datasets on lexical ambiguity \cite{haber-poesio-2021-patterns-polysemy, karidi-etal-2021-putting, schlechtweg2021dwug, erk2013measuring} could be augmented with sensorimotor judgments. Further, because the original RAW-C items were adapted from psycholinguistic studies \cite{trott-bergen-2021-raw}, those items might be skewed towards the phenomena those researchers were interested in; for example, it is possible that certain polysemous relationships (metaphor and metonymy) may be overrepresented.

A second, related problem is that the participants were sourced from an undergraduate population, which is likely non-representative of the broader population at large \cite{henrich2010most}. Similarly, the sentences themselves were hand-crafted, and thus do not reflect the full diversity of contexts these meanings might enjoy in naturalistic usage. Future work should attempt to ensure diversity in both the sample of annotators and the sentences under consideration.

Third, as others have noted \cite{bender-koller-2020-climbing, bisk-etal-2020-experience, tamari-etal-2020-language, borghi2019words}, \textit{grounding} goes beyond sensorimotor associations. Linguistic meaning is also grounded in social experience and interaction. Recent work has attempted to incorporate these social aspects of grounding, either by integrating social information into distributional models \cite{johns2021distributional} or simply by including more dimensions in the grounded feature representations \cite{binder2016toward}.

Finally, recent work has enjoyed some success in learning grounded feature vectors directly from LM representations, typically for words rated in isolation \cite{ turton-etal-2020-extrapolating, chersoni-etal-2020-automatic, utsumi2020exploring}. One question is whether contextualized embeddings, derived from a large LM such as BERT, are sensitive enough to capture the fine-grained distinctions that the CS Norms encode across sentential contexts for the same word. Of course, it is possible that the CS Norms dataset is simply too small to successfully augment a LM like BERT. However, the norms could also be used as a ``challenge set'', i.e., to \textit{evaluate} how much information about sensorimotor properties of a word are in principle derivable from an LM's representations. For example, the performance of an ungrounded model like BERT could be compared to recent multi-modal models \cite{zellers-etal-2021-piglet, zellers2021merlot}.

\section{Conclusion}

We have presented a novel resource: human judgments about the strength or salience of various sensorimotor features for 112 English words, each appearing in four distinct sentential contexts. This resource was extended from past work \cite{trott-bergen-2021-raw}, and thus also contains information about the relatedness \textit{between} sentential contexts for the same word. We provided several demonstrations of the dataset's utility, above and beyond judgments of these words in isolation \cite{lynott2019lancaster}, as well as large LMs such as BERT (see Section \ref{sec:utility}).

\section{Ethical Considerations}

All responses from human participants were anonymized before analyzing any data.

All participants provided informed consent, and were compensated in the form of class credit. The project was carried out with IRB approval. 

Finally, we have attempted to ensure dataset quality by: 1) removing responses from participants who failed bot checks; 2) removing participants whose inter-annotator agreement scores were more than two standard deviations below the average; and 3) collecting at least ten ratings per sentence, per judgment type, as in past work \cite{lynott2019lancaster}.

\section*{Acknowledgements}

\bibliography{anthology,custom}
\bibliographystyle{acl_natbib}

\appendix

\end{document}